\documentclass{article}

\usepackage{arxiv}

\usepackage[utf8]{inputenc} 
\usepackage[T1]{fontenc}    
\usepackage{hyperref}       
\usepackage{url}            
\usepackage{booktabs}       
\usepackage{amsfonts}       
\usepackage{nicefrac}       
\usepackage{microtype}      
\usepackage{lipsum}		
\usepackage{graphicx}
\usepackage[square,sort,comma,numbers]{natbib}
\usepackage{doi}

\title{Ray Based Distributed Autonomous Vehicle Research Platform}

\author{{\hspace{1mm}Derek Xu} \\
	SAAS Research \& Publications\\
	University of California, Berkeley\\
	Berkeley, California, USA\\
	\texttt{xzrderek@berkeley.edu} \\
}

\date{}

\renewcommand{\headeright}{Technical Report}
\renewcommand{\undertitle}{Technical Report}
\renewcommand{\shorttitle}{Ray-Carla Autonomous Vehicle Research Platform}

\hypersetup{
pdftitle={Ray-Carla Autonomous Vehicle Research Platform},
pdfsubject={q-bio.NC, q-bio.QM},
pdfauthor={Derek Xu},
pdfkeywords={Ray, Carla, Distributed Systems},
}

\begin{document}
\maketitle

\begin{abstract}
My project tackles the question of whether Ray can be used to quickly train autonomous vehicles using a simulator (Carla), and whether a platform robust enough for further research purposes can be built around it. Ray is an open-source framework that enables distributed machine learning applications. Distributed computing is a technique which parallelizes computational tasks, such as training a model, among many machines. Ray abstracts away the complex coordination of these machines, making it rapidly scalable. Carla is a vehicle simulator that generates data used to train a model. The bulk of the project was writing the training logic that Ray would use to train my distributed model. Imitation learning is the the best fit for autonomous vehicles. Imitation learning is an alternative to reinforcement learning and it works by trying to learn the optimal policy by imitating an “expert” (usually a human) given a set of demonstrations. A key deliverable for the project was showcasing my trained agent in a few benchmark tests, such as navigating a complex turn through traffic. Beyond that, the broader ambition was to develop a research platform where others could quickly train and run experiments on huge amounts of Carla vehicle data. Thus, my end product is not a single model, but a large-scale, open-source research platform (RayCarla) for autonomous vehicle researchers to utilize.

\end{abstract}


\section{Introduction}

\subsection{The Role of Ray in Distributed Machine Learning}

Machine learning is a critical tool for companies making data-driven decisions as well as researchers progressing humanity’s knowledge of AI. However, the demand for processing training data has long outpaced the increase in hardware computational power. As a result, there is a need to split, or distribute, the machine learning workload across many machines. This is what is known as distributed machine learning. The company Anyscale Inc. aims to do this with Ray. Ray is an open-source framework that allows users to quickly build and scale distributed machine learning applications by abstracting away the complex process of coordinating machines, CPU cores, and GPUs. Ray also includes a set of machine learning libraries, such as RLib (reinforcement learning) and Tune (hyperparameter tuning). The goal of this project is to investigate the effectiveness of Ray and whether it’s able to be used in real-world applications.

\subsection{Carla and Deep Imitative Models}

Carla is an open-source vehicle simulator often used by autonomous vehicle researchers. Carla also has a Python API that allows a user to control all aspects of the simulation, such as traffic, pedestrians, weather, sensors, and much more. Further, because our main purpose is not to develop the models that train self-driving cars but rather to make these processes more efficient with Ray, this project will use OATomobile and Deep Imitative Models (DIM). OATomobile is a library for autonomous driving research, developed by researchers from Oxford, who in turn used Deep Imitative Models in their work, developed by CMU and UC Berkeley researchers. The current implementation only supports a single worker, and cannot leverage distributed computing.

\subsection{Real World Applications of Distributed Machine Learning}

There are many real-world applications of distributed machine learning, but this project focuses on self-driving vehicles through imitation learning. Specifically, our end goal is not to show how one model was trained quickly, but rather to develop an open-source, large-scale research platform for autonomous vehicle researchers. Because of the nature of Ray and its ability to distribute the machine learning workload, this research platform is able to take in vast amounts of data and train a model extremely quickly. One can imagine the wide-scale effects and commercialization of such a platform. For instance, self-driving companies like Tesla [1] and Waymo [2] are able to quickly train models and run experiments on huge amounts of data.

\section{Background}

\subsection{Imitation Learning}

What is imitation learning? At a high level, imitation learning is a means for an agent to mimic some expert, typically a human, in a given task [3]. The agent learns to complete the task by mapping a series of observations to actions after a set of demonstrations by the expert. The agent effectively learns by “imitating”, hence the name. Imitation learning is usually considered as an alternative to reinforcement learning. In reinforcement learning, the agent learns from its own exploration from scratch, but in imitation learning, the model learns specific policies from demonstrations. Deep Imitative Models uses deep learning models to implement imitation learning for autonomous driving in Carla. My work’s purpose is to convert DIM to a high-performance research platform with distributed machine learning capabilities using Ray.

\subsection{DIM on Ray Architecture}

At a high level, this is what the architecture of the project will look like:

\begin{figure}[htbp]
\centerline{\includegraphics[scale = .42]{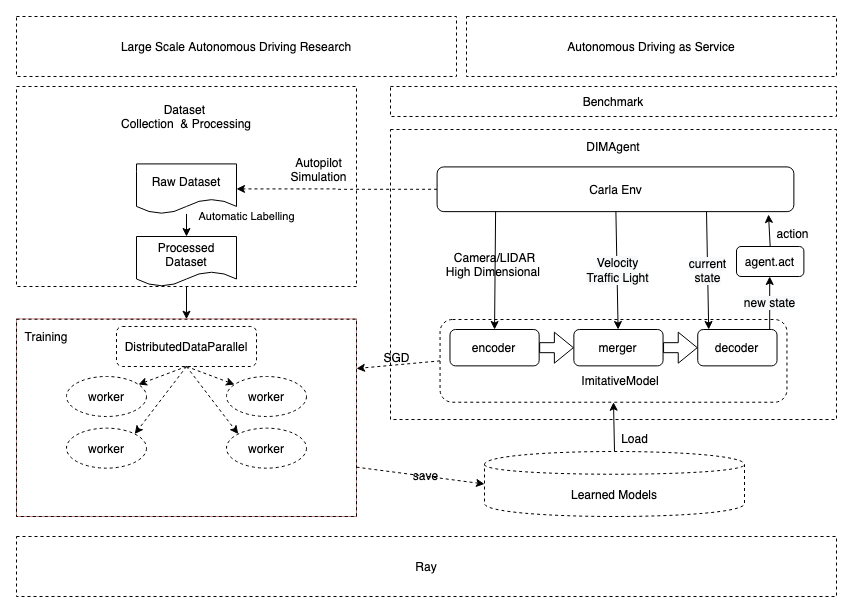}}
\caption{DIM on Ray Architecture}
\label{fig}
\end{figure}

DIM on Ray is composed of three subsystems: dataset collection and processing, training, and the agent. First, we need to collect the raw dataset using an autopilot agent from the Carla environment. Then, we need to automatically add labels to create a processed dataset. Next, we use this dataset to train the model and save it to a repository of learned models. The learned model is loaded into the DIMAgent, so that it can now predict how to drive in an environment such as a busy town. The details of the DIMAgent will be discussed more later. 

Further, here is how a user might use their learned model in the Carla environment after training it. Note that this is a simplification, but this provides an overall skeleton of how the model actually performs actions in a Carla town. 

\begin{figure}[htbp]
\centerline{\includegraphics[scale = .25]{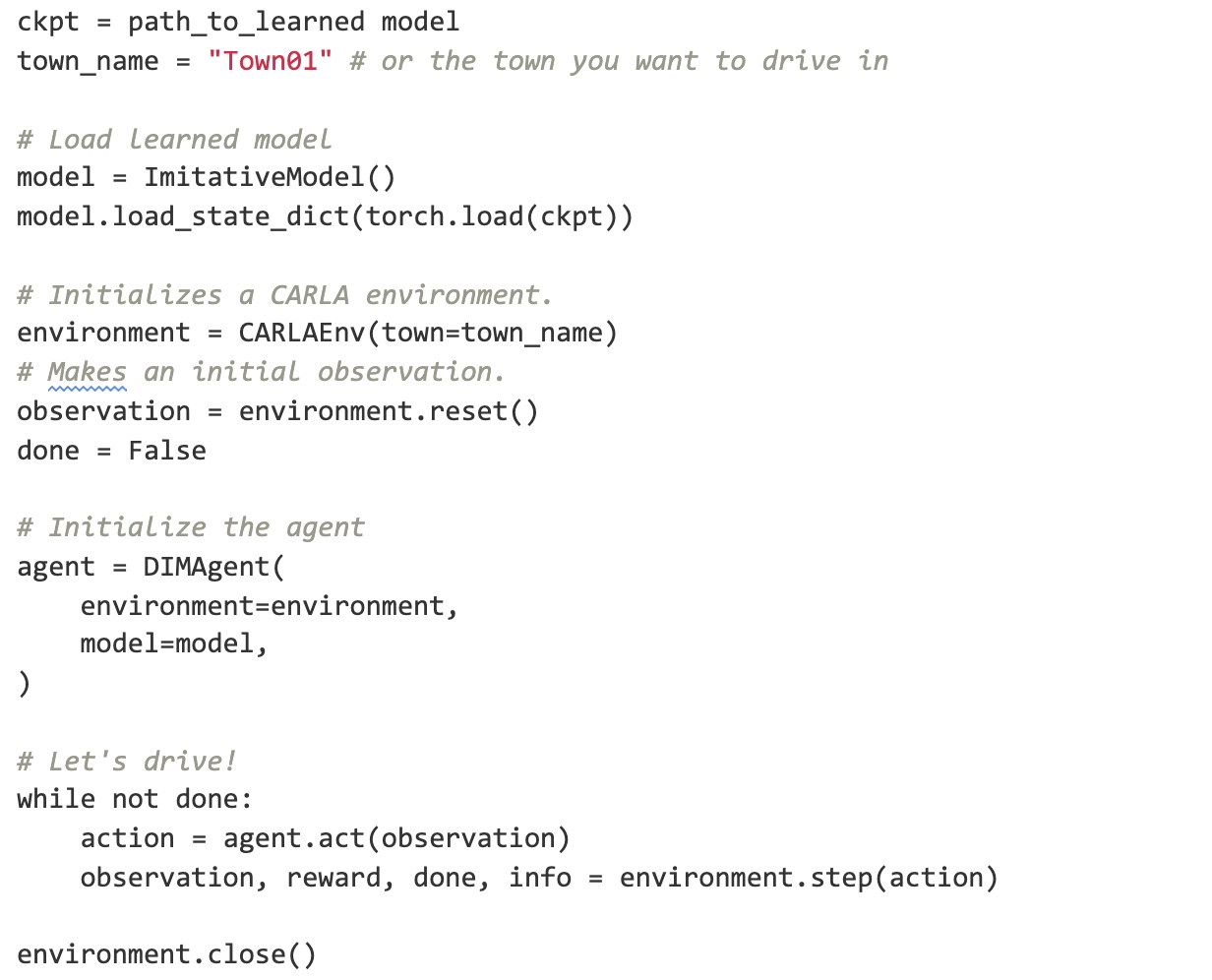}}
\label{fig}
\end{figure}

Our final goal is to implement all three subsystems on top of Ray. As of now, we only finished the training portion, which was chosen because it is the most time-consuming out of the three. 

\subsection{DIM Agent}
To formalize deep imitative models, we first introduce notation [4]. $s$ denotes the current state. $t$ represents time or the current time step. The state at time $t$ is represented as $s_t \in R^D, t = 0$, where $D = 2$. $\phi$ represents the agent’s observations. $\tau$ represents the number of past positions we condition on. $\chi$ represents a high-dimensional observation of the scene, such as LIDAR sensor data and camera images or both. $\lambda$ represents a low-dimensional observation of the scene, such as traffic light signal data and the vehicle’s velocity. We are able to featurize the LIDAR data as $\chi=R^{200 \times 200 \times 2}$. The agent has access to environment perception $\phi \leftarrow s_{-\tau:0}, \chi, \lambda$. Lastly, Carla provides ground truth $s_{-\tau:0}, \chi, \lambda$, meaning these are known variables that are certainly true.

Our purpose in formalizing these concepts is to find future time steps, such as $S := S_{1:T}R^{T \times D}$. We do this in order to fit an imitative model, 

$$q(S_{1:T} | \phi) = \prod_{t=1}^T q(S_t|S_{1:t-1}, \phi)$$

to a dataset of expert trajectories

$$D=(s^i, \phi^i)_{i=1}^N$$

which is data simulated from Carla. The goal is to train $q(S|\phi)$ to generate trajectories that an expert would generate. While this is a simplification, this encapsulates the process of deep imitative learning.

The entire process can be summarized by this figure [4]:

\begin{figure}[htbp]
\centerline{\includegraphics[scale = .12]{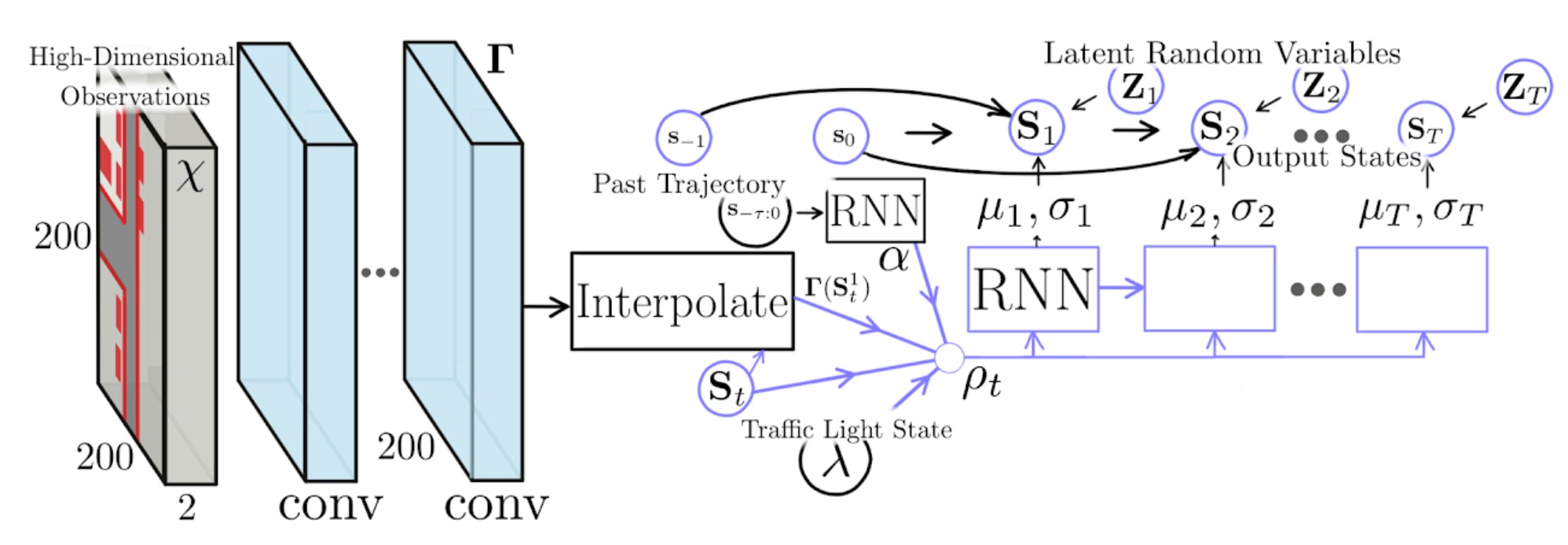}}
\label{fig}
\end{figure}

\subsubsection{State}

In our autonomous driving application, we model the agent’s state at time $t$ as $s_t \in R^2$ and $s_t$ represents our agent’s location on the ground plane. $s_0$ is the current location, $s_{-\tau:0}$ is the past trajectory of waypoints, and $s_{1:T}$ is the future trajectory of waypoints, and what our model is predicting.

\subsubsection{Observation Space}

It’s important to establish what the observation space looks like for our model. In code, our observation space looks like this:

\begin{figure}[htbp]
\centerline{\includegraphics[scale = .25]{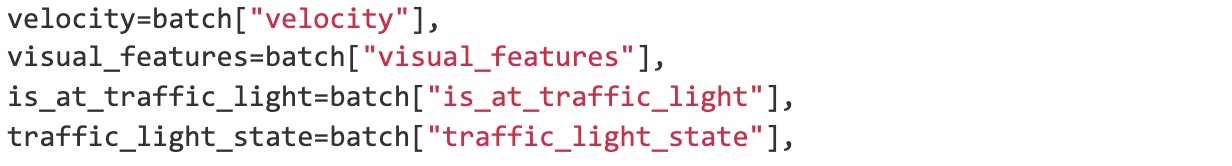}}
\label{fig}
\end{figure}

As one can see, the four observations that can be made are velocity, visual\_features, is\_at\_traffic\_light, and traffic\_light\_state. This means that in the Carla simulator, the model is able to make observations on those four features, and are either high-dimensional or low-dimensional type, and used to determine the eventual state of the model. 

\subsubsection{Action Space}

The action space is defined as such in code:

\begin{figure}[htbp]
\centerline{\includegraphics[scale = .25]{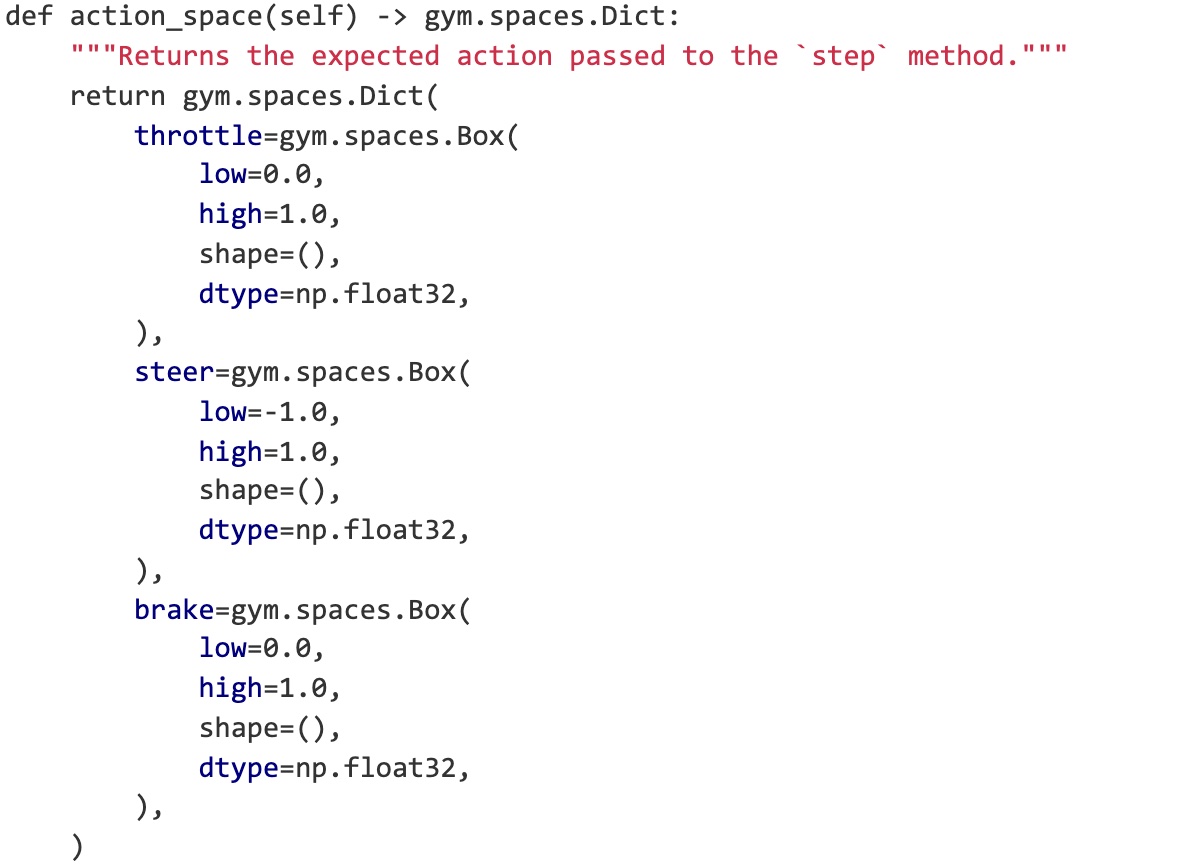}}
\label{fig}
\end{figure}

The agent is responsible for translating the predicted waypoint into actions (throttle, steering, and brake). Thus, the three actions the model is able to take are throttle (driving forward or back), steer, or brake. 

\subsubsection{Model}

Moving on to how the actual model is defined in pseudocode: 

\begin{figure}[htbp]
\centerline{\includegraphics[scale = .25]{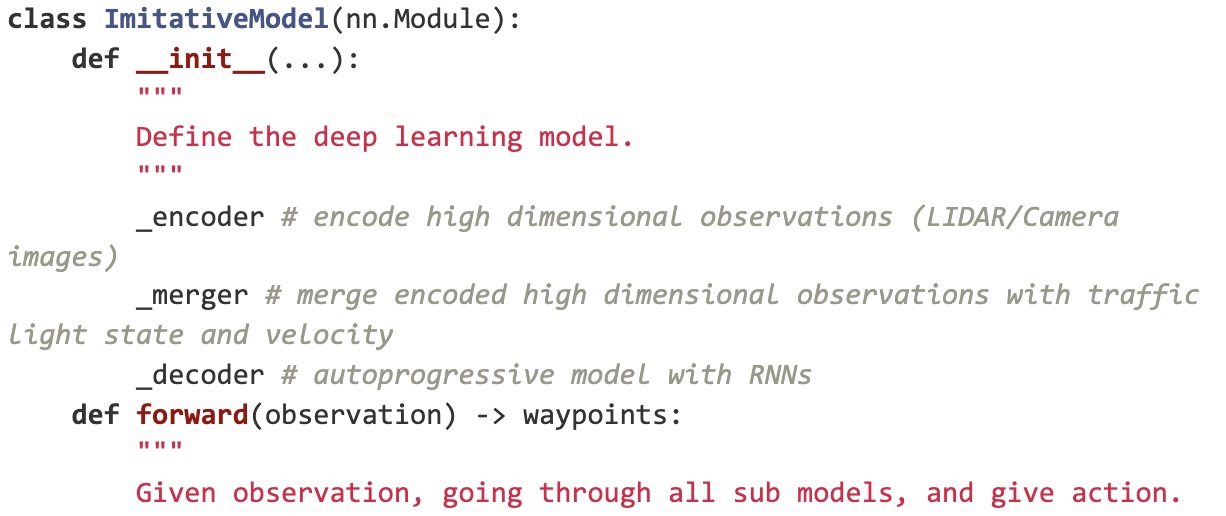}}
\label{fig}
\end{figure}

In order to create a DIM, we use three submodels: encoder, merger, and decoder [4]. The encoder takes in high dimensional observation data such as LIDAR or camera images and encodes them to a vector. The merger takes the encoded high dimensional features and merges them with low dimensional observations such as traffic light state and velocity. Finally, the decoder uses an AutoregressiveFlow model with RNNs (recurrent neural networks) and takes these merged values and current state to find the next state. The code snippet above is simplified pseudocode for the user to more easily understand, but the full implementation is below in section Section III, A.5 Model Implementation.

\subsubsection{Agent}

Agent is an intelligent “agent” which provides the method act(). The agent gets the current state (location) and observations from the Carla environment, then queries the model to get the prediction of the next state (new waypoints in desired trajectory), and finally turns these waypoints into actions. The user invokes agent.act(observation) to get an action, which invokes env.step(action):

\begin{figure}[htbp]
\centerline{\includegraphics[scale = .25]{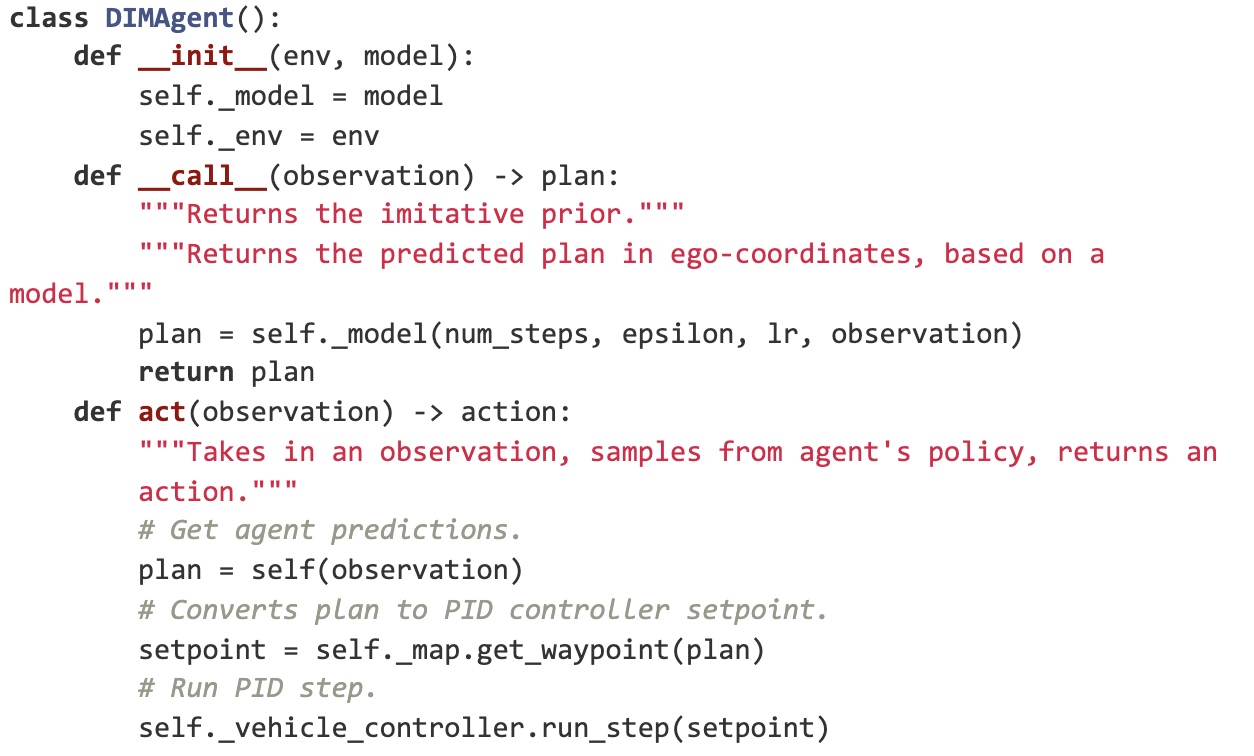}}
\label{fig}
\end{figure}

\section{Implementation}

\subsection{Ray Train}

Most of our implementation will focus on converting the previous training logic to training logic using a specific library of Ray called Ray Train [7]. This was accomplished by referencing their quick start guide and documentation and porting our code to Ray Train. 

First, let’s take a look at what exactly we are training on. When we want to train a model, we start with 200001 training samples and 40001 validation samples. We separate the samples into batches. We have 512 samples per batch for training and $512 \times 5$ samples per batch for validation. Next, we split our training sequence into 200 epochs, or episodes of training. For each epoch, we train $200001 / 512 = 391$ batches. For each epoch, we validate $40001/(512 \times 5) = 16$ batches. In each epoch, we will train using all 200001 training samples, but randomize the order every time. With each epoch, we hope for a decrease in loss. We will take a look side by side at exactly what code was changed. For all code snippets below, the highlighted lines are modified or added.

\subsubsection{Ray Init}

We must instantiate a Ray Train Trainer, which is the primary object used to manage state and execute training. We’re able to choose a specific backend such as “torch”, “tensorflow”, or “horovod”, but we choose to use torch in this case. 

\begin{figure}[htbp]
\centerline{\includegraphics[scale = .25]{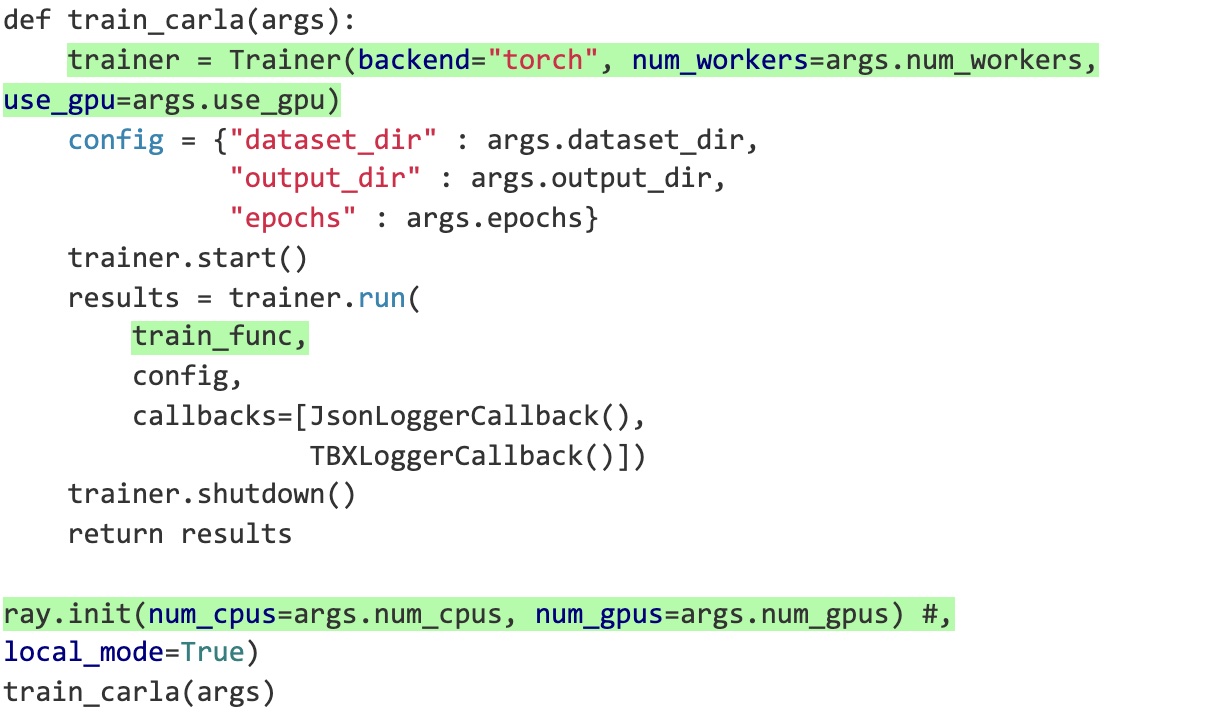}}
\label{fig}
\end{figure}

\subsubsection{Ray Task}

We must change the name of our single-worker training function because we’re no longer executing it just once, but we instead want a distributed multi-worker training function. Therefore, we need it to be more general than “main”. We change the name to “train\_func”.

\begin{figure}[htbp]
\centerline{\includegraphics[scale = .25]{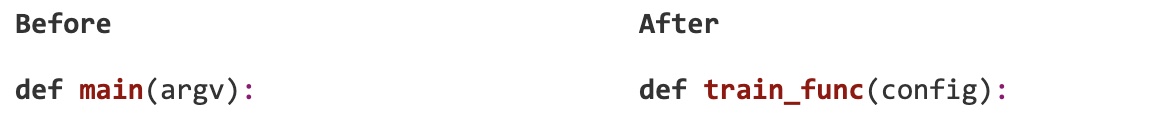}}
\label{fig}
\end{figure}

\subsubsection{GPU Setup}

Because we are using GPUs, we must properly set up our CUDA devices. If one is not using GPUs, they can skip this step. 

\begin{figure}[htbp]
\centerline{\includegraphics[scale = .25]{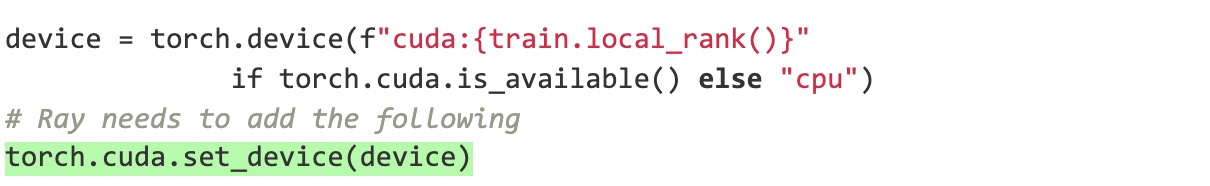}}
\label{fig}
\end{figure}

\subsubsection{Model Wrapping}

Next, we need to wrap our model in the DistributedDataParallel container, because this allows the input to be parallelized across the various workers. 

\begin{figure}[htbp]
\centerline{\includegraphics[scale = .26]{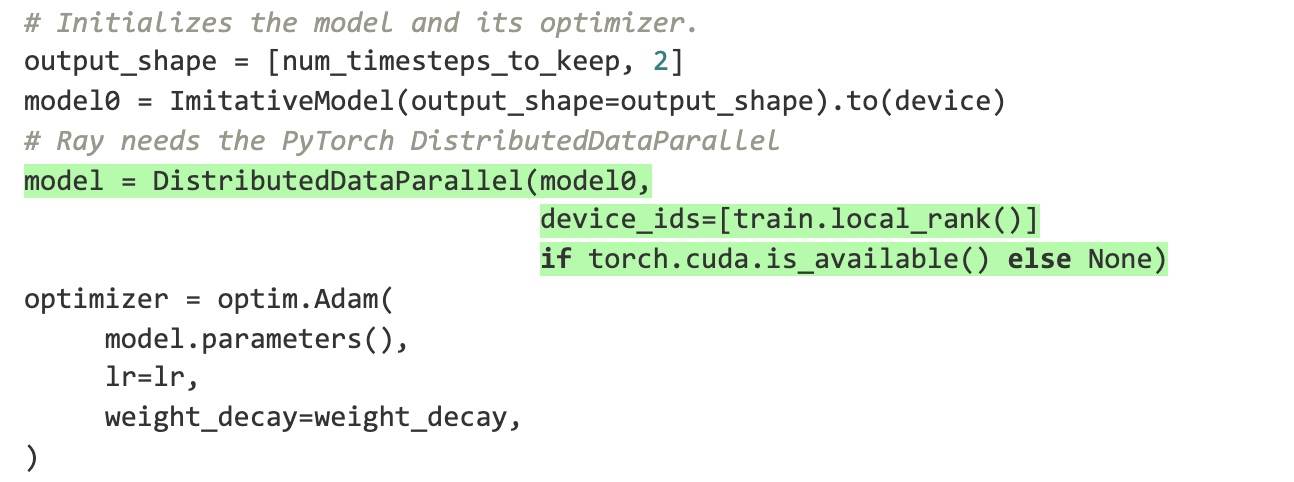}}
\label{fig}
\end{figure}

\subsubsection{DataLoader}

We also need to update our DataLoader using a DistributedSampler container, which splits data across the workers. Thus, the training and validation samples will be split across the separate machines accordingly. This ensures that each worker only trains on a subset of the data. We are currently working on further improving the efficiency of distributed data ingest with Ray Dataset. It provides benefits such as pipelining and automatic locality-aware sharding.

\begin{figure}[htbp]
\centerline{\includegraphics[scale = .23]{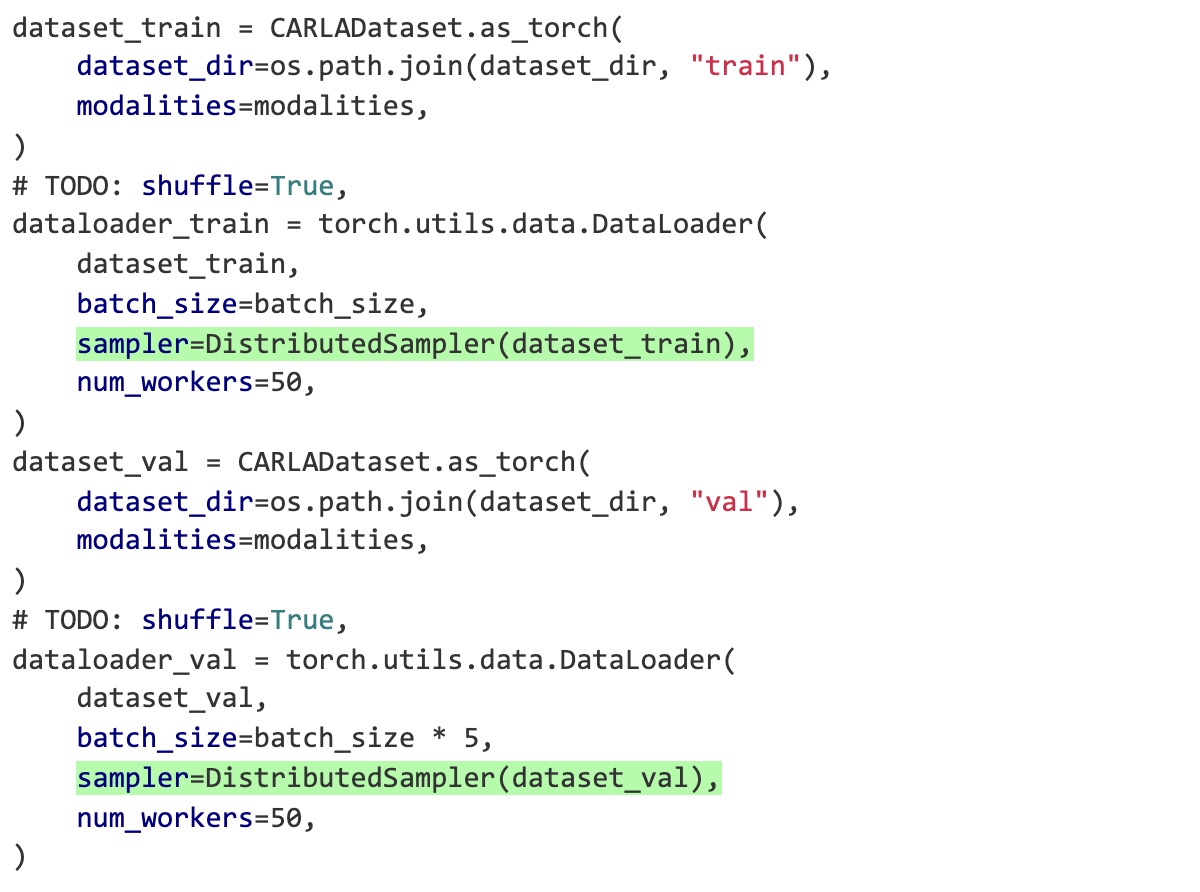}}
\label{fig}
\end{figure}

\subsubsection{Training Logic}

train\_step:

\begin{figure}[htbp]
\centerline{\includegraphics[scale = .23]{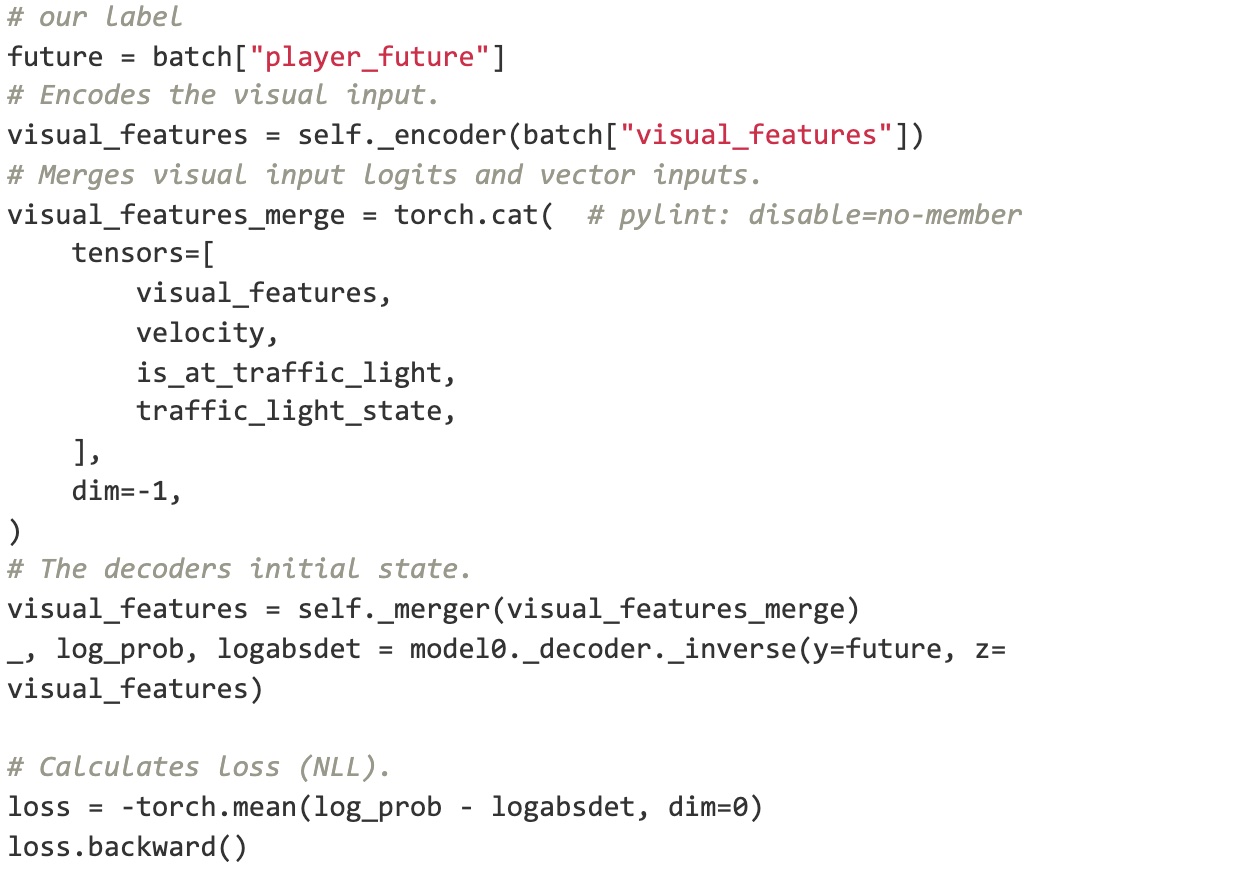}}
\label{fig}
\end{figure}

train\_epoch:

\begin{figure}[htbp]
\centerline{\includegraphics[scale = .21]{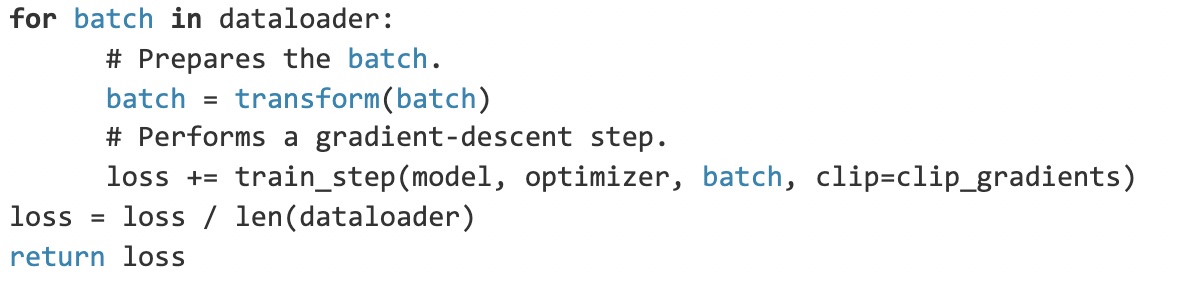}}
\label{fig}
\end{figure}

These two code snippets show the actual process of training the model. We did not need to change the code very much to port it to Ray Train, which is another key advantage of Ray.

\subsection{Why is Ray Necessary?}

From the code snippets above, it appears as though the changes are very simple, and that Ray is not doing much, but this is not the case. Ray is helping handle a lot of the complex distributed issues. For example, if we have 10 machines, and each machine has 16 workers in parallel, and each worker is checkpointing the model, what is the desired behavior? We don’t want all 16 workers to be checkpointing because that’d be redundant, but we still need checkpoints made. Ray handles it so that the 0th ranked worker does the checkpointing on each machine. This is just one example of why we use Ray to help with these distributed issues.

\section{Results}

\subsection{Training Performance}

Without Ray, training the imitation model over 200 epochs using 1 GPU and 1 CPU took $\textbf{37589 seconds (10.5 hours)}$. With Ray, training the imitation model over 200 epochs using 8 GPUs (25 epochs each) and 8 CPUs took $\textbf{2278 seconds (38 minutes)}$. Lastly, if we wanted to train 200 epochs per GPU, it took 18200 seconds (3 hours) to finish 1600 epochs of training using 8 GPUs and 8 CPUs. The 1600 epoch model is not as relevant when comparing performance between Ray and without Ray.

\subsubsection{Hardware and System Monitoring}

While training, we are able to see the distribution of GPU and CPU load using the command htop and CUDA command nvidia-smi.

\begin{flushleft}
\emph{Without Ray:}
\end{flushleft}

\begin{figure}[htbp]
\centerline{\includegraphics[scale = .145]{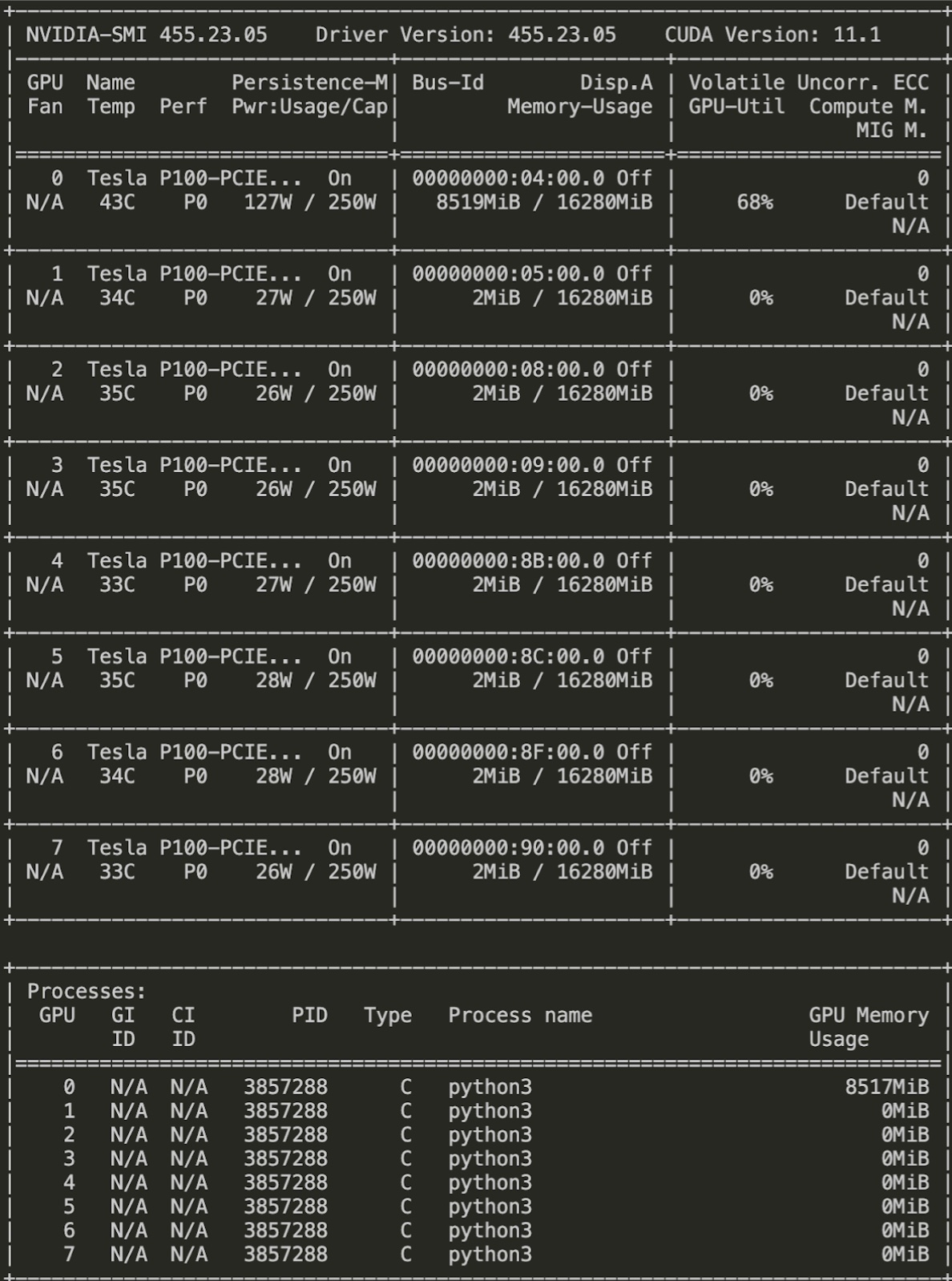}}
\label{fig}
\end{figure}

\begin{figure}[htbp]
\centerline{\includegraphics[scale = .22]{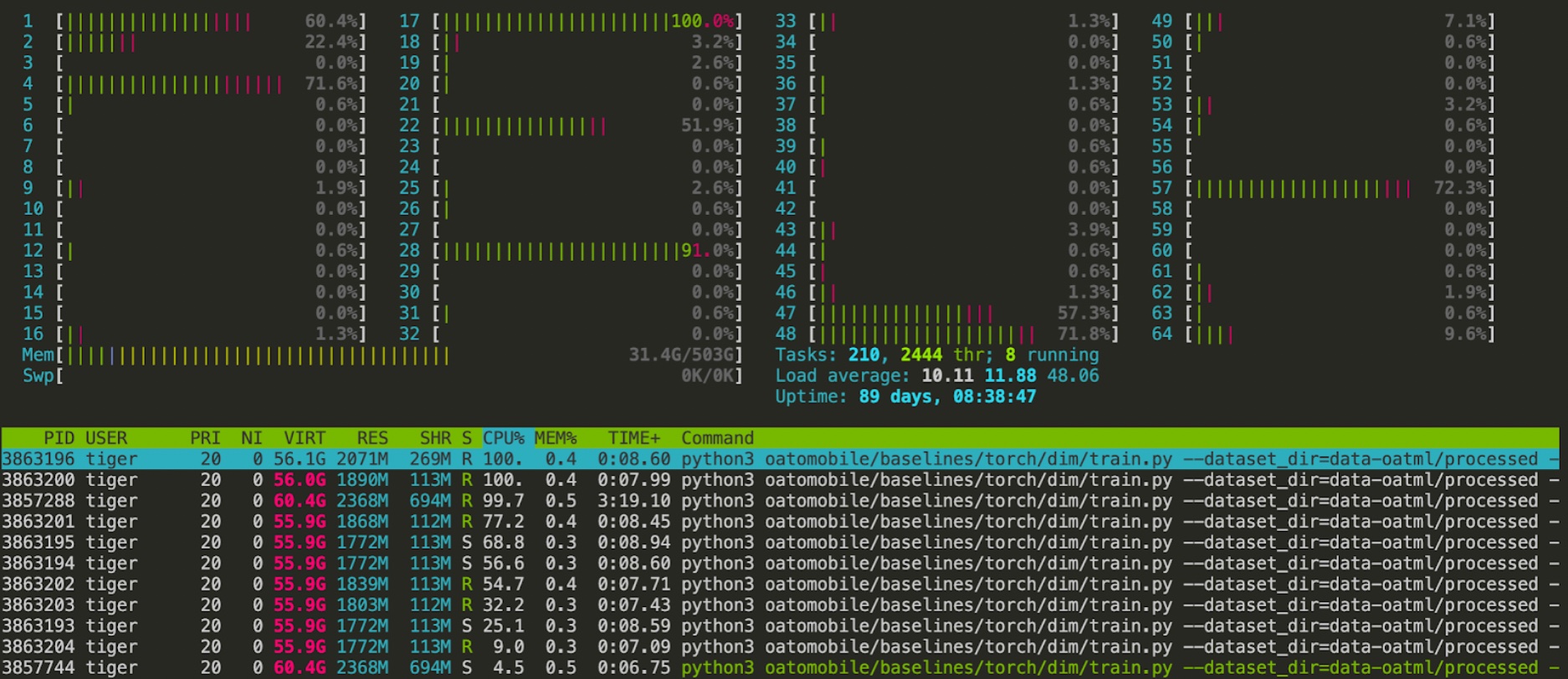}}
\caption{htop and nvidia-smi commands showing 1 GPU under max load without Ray.}
\label{fig}
\end{figure}

In the first image, you can see that only one GPU is busy while the other 7 are idle. Each worker uses 50 threads, and the second image shows a number of cores busy. Only a few cores are busy because the rest are idling, waiting for the disk IO.

\begin{flushleft}
\emph{With Ray:}
\end{flushleft}

\begin{figure}[htbp]
\centerline{\includegraphics[scale = .23]{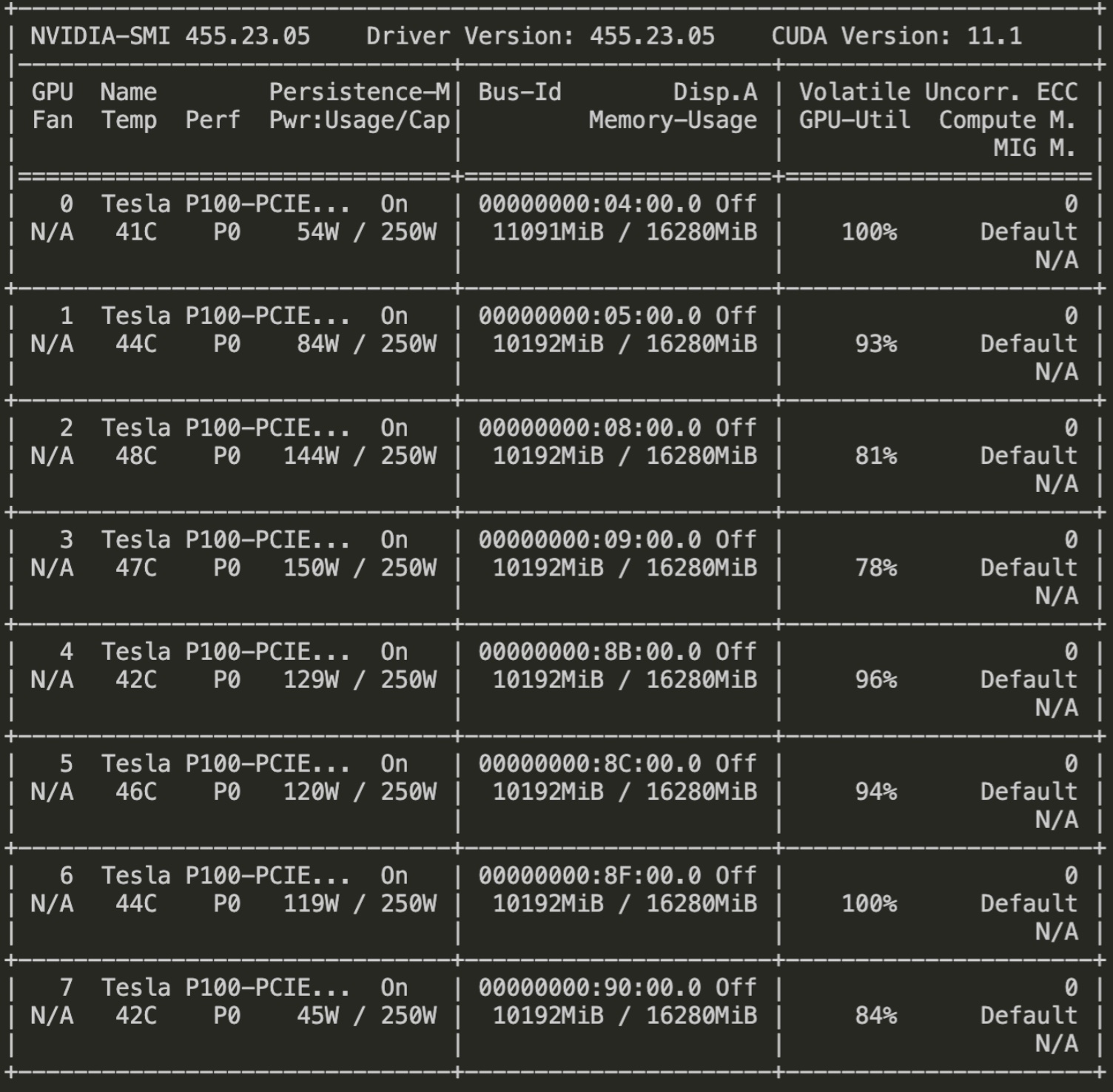}}
\label{fig}
\end{figure}

\begin{figure}[htbp]
\centerline{\includegraphics[scale = .16]{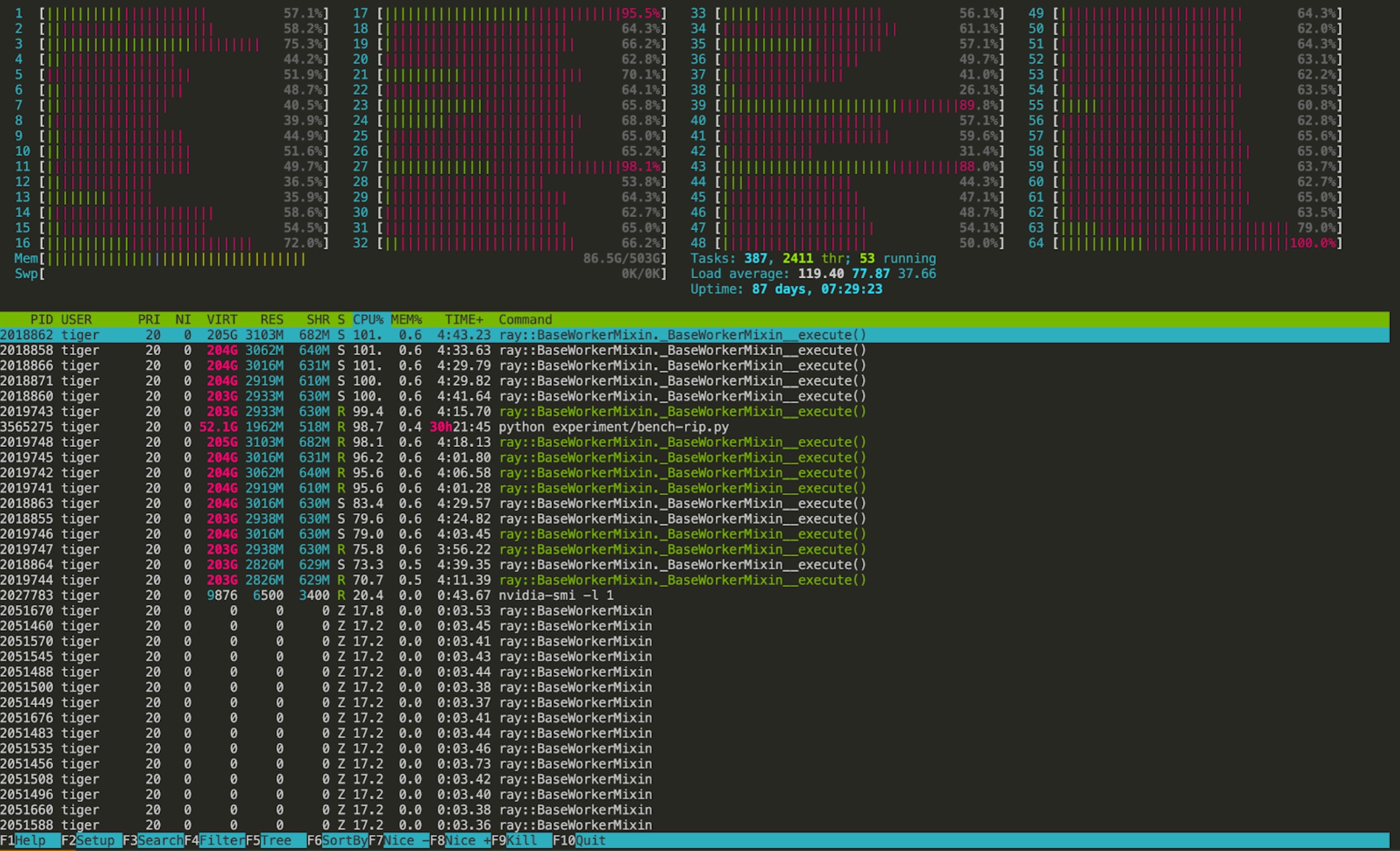}}
\caption{htop and nvidia-smi commands showing all CPUs, all GPUs under max load with Ray.}
\label{fig}
\end{figure}

In contrast with the images above, all 8 GPUs and CPU threads are busy, and none are idle. This means all our resources are being used to their max. This is why we are able to reduce 10 hours to 40 minutes.

\subsection{Model Performance}

Using TensorBoard, we can see a loss pattern on all three models we trained. In summary, we used Distributed Data-Parallel to train on an 8 GPU machine. The y-axis represents the negative log of the probability of future states given the observation:

$$q(S_{1:T} | \phi) = \prod_{t=1}^T q(S_t|S_{1:t-1}, \phi)$$

\subsubsection{Without Ray, 1 Worker, 200 Epochs}

In this model, we used 1 GPU to train 200 epochs of the model.

\begin{figure}[htbp]
\centerline{\includegraphics[scale = .17]{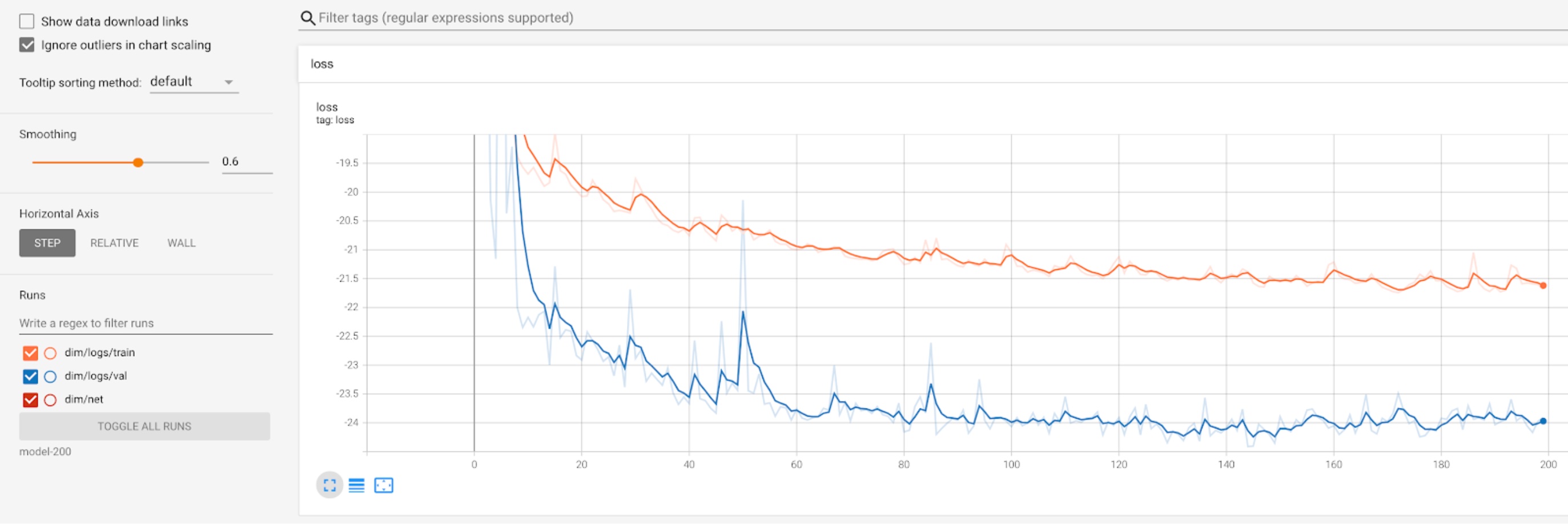}}
\caption{TensorBoard line graph showing loss over epoch number for 200 epochs trained across 1 GPU and 1 CPU.}
\label{fig}
\end{figure}

\subsubsection{With Ray, 8 GPUs and 8 CPUs, 200 Epochs Total}

In this model, we used 8 GPUs to train the model, and the work was distributed so that each GPU trained for 25 epochs, which is why the x-axis of the graph in the two images below only goes from 0 to 25.

\begin{figure}[htbp]
\centerline{\includegraphics[scale = .17]{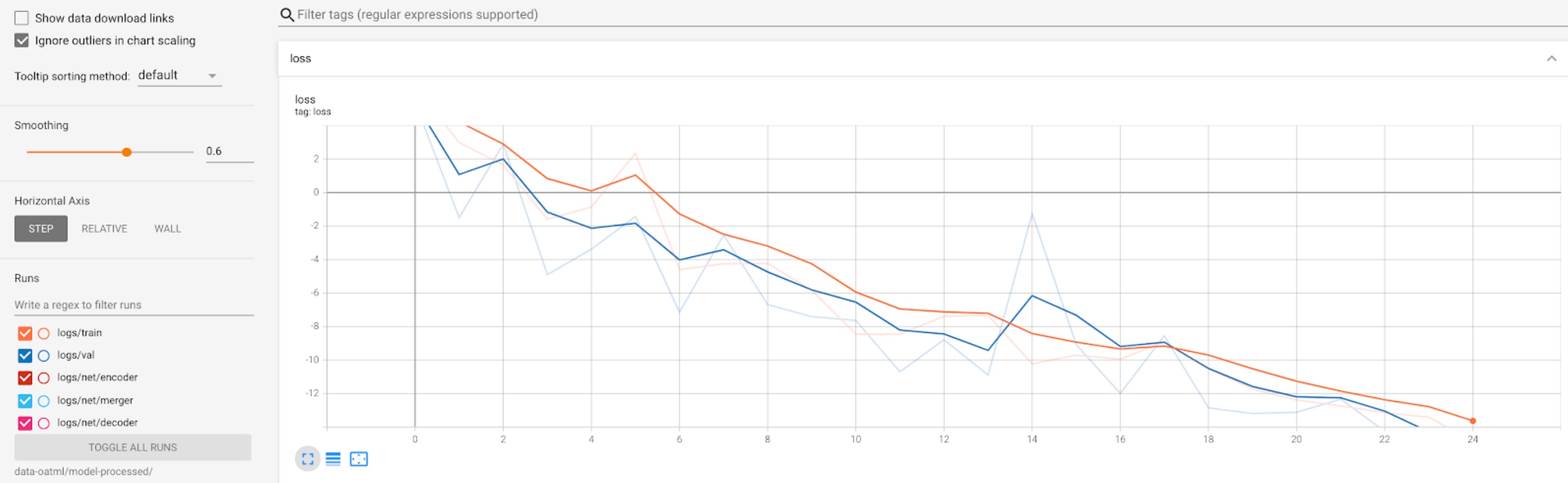}}
\caption{TensorBoard line graph showing loss over epoch number for 200 epochs trained across 8 GPUs and 8 CPUs.}
\label{fig}
\end{figure}

\subsubsection{With Ray, 8 GPUs and 8 CPUs, 1600 Epochs Total}

In this model, we used 8 GPUs to train the model, and the work was distributed so that each GPU trained for 200 epochs, which is why the x-axis of the graph in the two images below goes from 0 to 200.

\begin{figure}[htbp]
\centerline{\includegraphics[scale = .17]{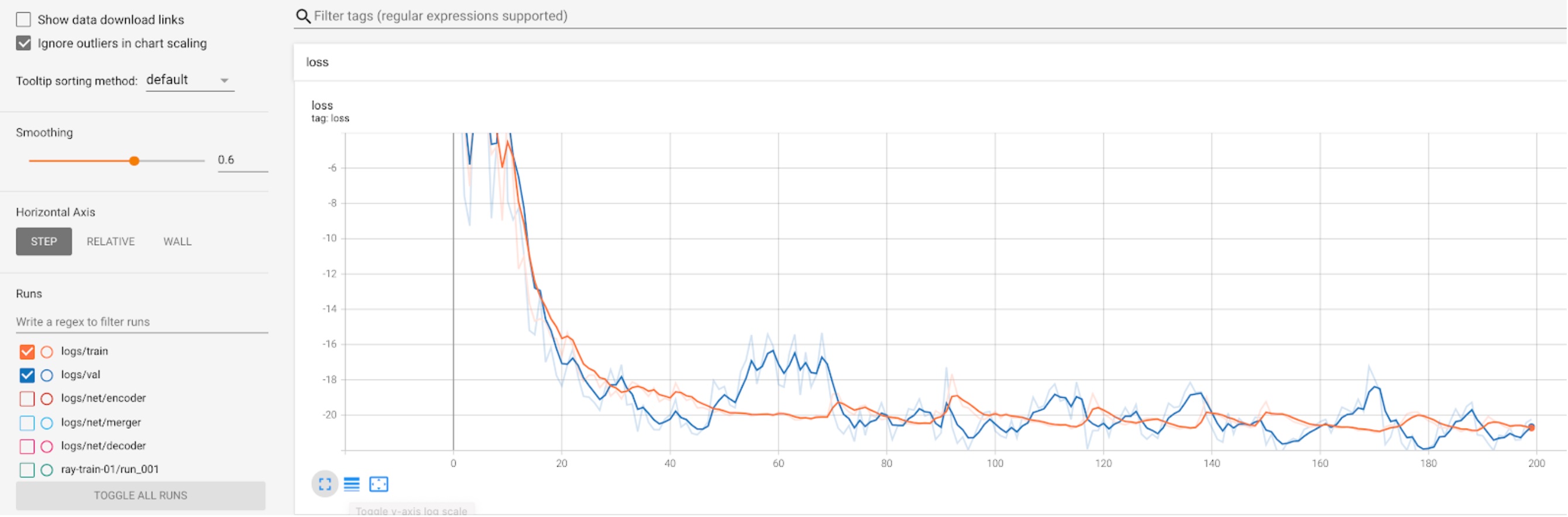}}
\caption{TensorBoard line graph showing loss over epoch number for 1600 epochs trained across 8 GPUs and 8 CPUs.}
\label{fig}
\end{figure}

\subsection{Benchmark Results}

In order to actually see if our model was successful, we can leverage the CarNovel benchmark library in OATomobile [6]. In their library, there are 27 test cases separated into the following categories: 7 AbnormalTurns, 11 BusyTown, 4 Hills, 5 Roundabouts. For example, AbnormalTurns 0 has the following configuration in the form of a JSON file:

\begin{figure}[htbp]
\centerline{\includegraphics[scale = .25]{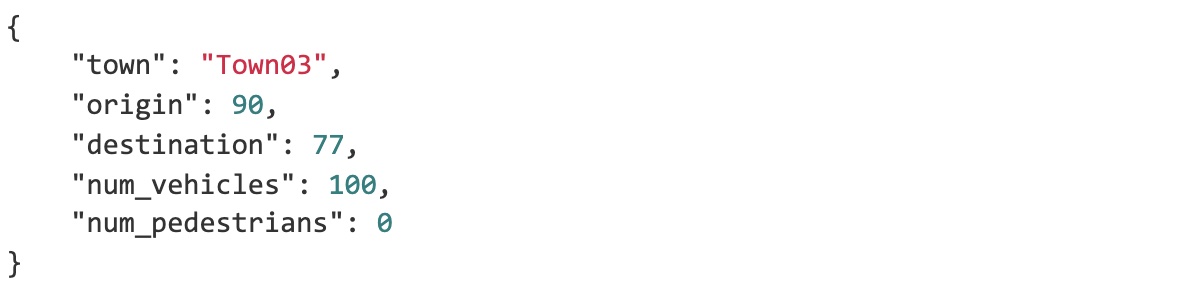}}
\label{fig}
\end{figure}

This means that in Carla Town 3, we drive from position 90 to position 77, and there are 100 vehicles and 0 pedestrians simulated. This trajectory is what we are trying to accomplish:

\begin{figure}[htbp]
\centerline{\includegraphics[scale = .45]{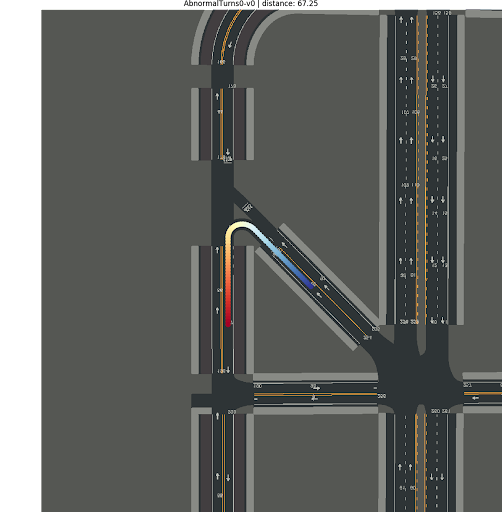}}
\caption{Goal trajectory depicted of an abnormal turn, car travels from red to blue.}
\label{fig}
\end{figure}

Here’s a video of what our model produced:

\begin{figure}[htbp]
\centerline{\includegraphics[scale = .75]{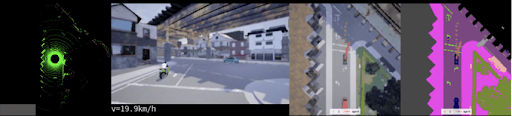}}
\centerline{\includegraphics[scale = .75]{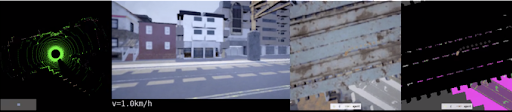}}
\centerline{\includegraphics[scale = .75]{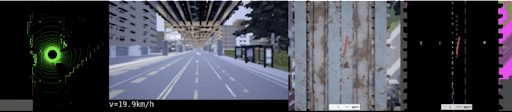}}
\caption{Three images showing the car before the abnormal turn, during the turn, and after the turn. From left to right: LIDAR, first-person camera view, bird-eye view, bird-eye view in RGB. Full video: https://youtu.be/G7O6AdhjYnk.}
\label{fig}
\end{figure}

As you can see, it accomplished the desired trajectory without colliding with any other vehicles. We can check in the test’s output that there were no collisions or lane invasions in a distance of 60.446 and 473 steps.

\begin{figure}[htbp]
\centerline{\includegraphics[scale = .7]{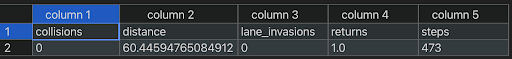}}
\caption{CSV file showing the results of our model’s benchmark test.}
\label{fig}
\end{figure}

Here’s another example of our model traversing through a roundabout:

\begin{figure}[htbp]
\centerline{\includegraphics[scale = .4]{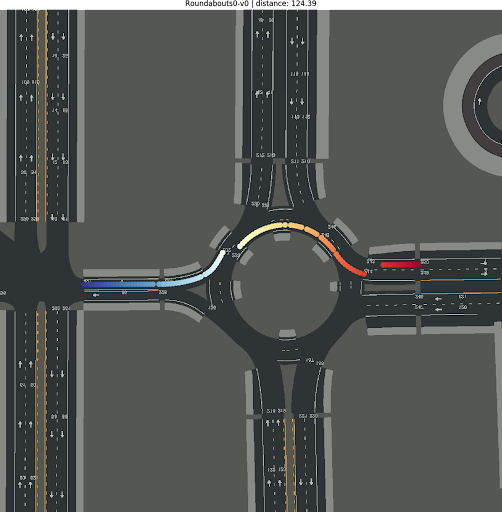}}
\caption{Goal trajectory depicted of a roundabout turn, car travels from red to blue.}
\label{fig}
\end{figure}

\begin{figure}[htbp]
\centerline{\includegraphics[scale = .8]{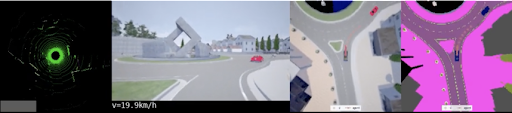}}
\centerline{\includegraphics[scale = .8]{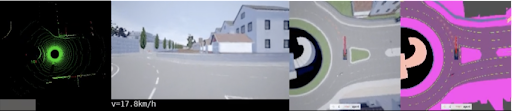}}
\centerline{\includegraphics[scale = .8]{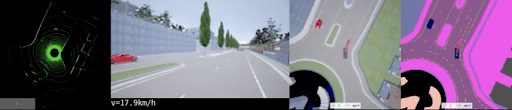}}
\label{fig}
\caption{Three images showing the car before the abnormal turn, during the turn, and after the turn. From left to right: LIDAR, first-person camera view, bird-eye view, bird-eye view in RGB.  Full video: https://youtu.be/L-ulcJDFLbY.}
\label{fig}
\end{figure}

\section{Conclusion}

\subsection{Improvements Made}

Right away, we notice promising results in the differences in speed while training the imitative models between using Ray and without Ray. Using Ray, the model trained significantly faster. Training 25 epochs on 8 GPUs (200 total) is 16.5 times faster than just training 200 epochs on one GPU without Ray (10.5 hours to 38 minutes). Training 200 epochs over 8 GPUs (1600 total) is still over 3 times faster than just 200 epochs on one GPU without Ray (10.5 hours to 3 hours). Further, we can see that the videos show our model driving along the desired trajectory correctly and the model was able to match the expert. If we were to have more nodes, say a total of 10 nodes, our Ray model theoretically would train at a linearly faster time, which would be 4 minutes (40/10). This would greatly increase the speed at which researchers can train and experiment on autonomous vehicle models.

One thing to note is that we expected the 1600 epoch training time to take around the same amount of time as without Ray (10.5 hours), but it was actually three times faster at 3 hours. One possible reason for this is that we have 8 times more IO workers when using Ray. If this is the case, this would confirm that data loading takes a significant percentage of the total training time and we are working on implementing Ray Dataset to make data ingestion more efficient. We expect its ability to pipeline IO and training to cut the time down even further. 
\subsection{Observed Issues and Future Work}

First, the most glaring issue is an issue with model performance. In Figure 4, we can see that the loss between the training set and the validation set is around -22. This was the model trained without Ray on 200 epochs using 1 GPU. On the other hand, in Figure 5, we can see that the loss between the training set and the validation set is around -13. This was the model trained using Ray with a total of 200 epochs across 8 GPUs. The performance of the latter model is significantly worse than the former model. We originally expected the performance of these both to be about the same, and don’t have a compelling reason for why this is the case. One potential reason is that the DistributedSampler is causing the problem. Or, it’s a bug within Ray. This requires further investigation. However, with the 1600 epoch model, in Figure 6, we can see that the loss approaches -22 at around 70 epochs, with similar performance as the model trained without Ray. Thus, theoretically, we could train a model using Ray on 8 GPUs with 70 epochs each and achieve the equivalent performance. This would take approximately 1 hour because it’s about 1/3 the epochs of the 3-hour training time, which is still one-tenth the training time of the non-Ray model.

A second issue is that we observed IO to be a potential bottleneck in the training process. As mentioned above, we are planning to investigate further into how Ray Dataset [7] can further bring the time down.

A third issue is that the CPU and GPU utilization was fluctuating throughout the training process. This means that our resources were not being maximized and could be made even more efficient. The further coordination of GPU and CPU operations along with Ray Dataset pipelining would help solve this problem.

In terms of development environment issues, we struggled with Carla and Ray requiring two different versions of Python (3.5 for Carla 0.9.6 and 3.7 for Ray). As a result, we had to always have two Python environments. We should port OATomobile to Carla 0.9.13+ to resolve this.

As mentioned in the high-level architecture of the entire project, we currently only have training enabled with Ray. In the future, we hope to enable dataset collection and inference/benchmark with Ray. End-to-end research would greatly benefit from this.

Keeping in mind our eventual goal of creating a Tesla-level autonomous driving research platform, we will need to look into more advanced topics such as Monte Carlo Tree Search [9] and GFlowNet [10].

\section*{Acknowledgment}

I’d like to thank Postdoc Nick Rhinehart of CMU, Rowan McAllister of UC Berkeley, and Professor Sergey Levine of UC Berkeley for their work in Deep Imitative Models. We’d also like to thank Angelos Filos of Oxford group OATML for his work on OATomobile.

\end{document}